\documentclass{article}

\usepackage[preprint]{neurips_2025}

\usepackage[utf8]{inputenc} 
\usepackage[T1]{fontenc}    
\usepackage{hyperref}       
\usepackage{url}            
\usepackage{booktabs}       
\usepackage{amsfonts}       
\usepackage{nicefrac}       
\usepackage{microtype}      
\usepackage{xcolor}         
\usepackage{multirow}
\usepackage{enumitem}

\usepackage{amsmath}
\usepackage{tikz}

\title{Enhancing Complex Instruction Following for Large Language Models with Mixture-of-Contexts Fine-tuning}

\author{%
  \textbf{Yuheng~Lu\textsuperscript{1}},
  \textbf{ZiMeng~Bai\textsuperscript{1}},
  \textbf{Caixia~Yuan\textsuperscript{1}},
  \textbf{Huixing~Jiang\textsuperscript{2}},
  \textbf{Xiaojie~Wang\textsuperscript{1,}}\thanks{Corresponding author\makeatother}
  \\
  \textsuperscript{1}School of Artificial Intelligence, Beijing University of Posts and Telecommunications \\
  \textsuperscript{2}LI Auto Inc.
  \\
  \{yuheng.lu, zimengbai, yuancx, xjwang\}@bupt.edu.cn \\
  jianghuixing@lixiang.com%
}

\begin{document}

\maketitle

\begin{abstract}
  Large language models (LLMs) exhibit remarkable capabilities in handling natural language tasks;
  however, they may struggle to consistently follow complex instructions including those involve multiple constraints.
  Post-training LLMs using supervised fine-tuning (SFT) is a standard approach to improve their ability to follow instructions.
  In addressing complex instruction following, existing efforts primarily focus on data-driven methods that synthesize complex instruction-output pairs for SFT.
  However, insufficient attention allocated to crucial sub-contexts may reduce the effectiveness of SFT.
  In this work, we propose transforming sequentially structured input instruction into multiple parallel instructions containing sub-contexts.
  To support processing this multi-input, we propose MISO (Multi-Input Single-Output), an extension to currently dominant decoder-only transformer-based LLMs.
  MISO introduces a mixture-of-contexts paradigm that jointly considers the overall instruction-output alignment and the influence of individual sub-contexts to enhance SFT effectiveness.
  We apply MISO fine-tuning to complex instruction-following datasets and evaluate it with standard LLM inference.
  Empirical results demonstrate the superiority of MISO as a fine-tuning method for LLMs,
  both in terms of effectiveness in complex instruction-following scenarios and its potential for training efficiency.
\end{abstract}

\section{Introduction}

Large language models (LLMs) exhibit remarkable capabilities in processing natural language tasks.
Among these, instruction following, which involves adhering to user intentions to generate helpful responses, plays a crucial role for practical usage.
A standard approach to enhance this capability is to perform supervised fine-tuning (SFT) using instruction-response pairs.

\begin{figure*}[t]
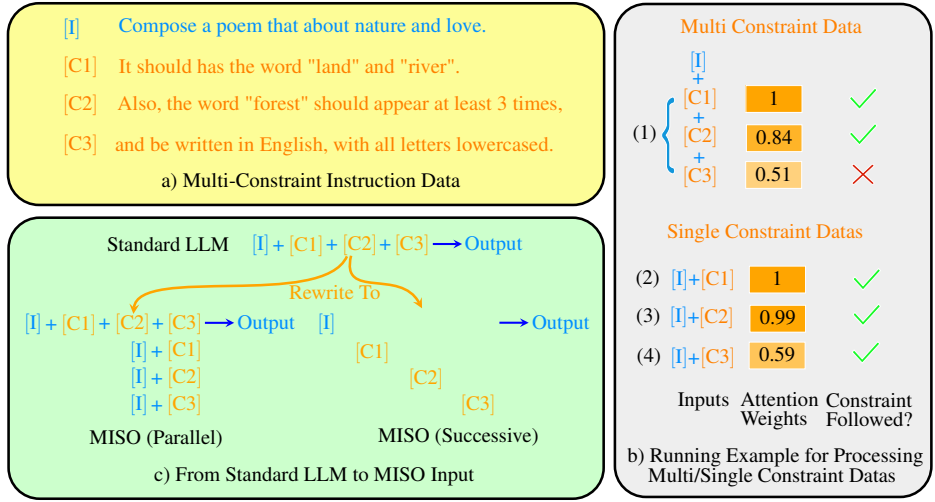

  \centering
  \resizebox{0.9\textwidth}{!}{
    \begingroup
    \large
\ifx\du\undefined
  \newlength{\du}
\fi
\setlength{\du}{15\unitlength}


    \endgroup
  }
  \caption{a) Example multi-constraint instruction following data; b) Attention allocation and whether instruction following success for multi-constraint input and corresponding single-constraint datas;
    c) Modeling structure for stardard LLM and MISO.}
  \label{fig:forget}
\end{figure*}

However, model performance tends to degrade when handling complex instructions, including those involving multiple constraints \citep{wenBenchmarkingComplexInstructionFollowing2024a}.
This limitation hinders the practical deployment of LLMs in real-world applications.
To address this challenge, recent works primarily focus on synthesizing complex instructions with LLMs to create corresponding instruction-output data \citep{xuWizardLMEmpoweringLarge2023, qiConstraintBacktranslationImproves2024, liRuleRImprovingLLM2024}.
Supervised fine-tuning (SFT) with standard language modeling loss is generally conducted for fine-tuning LLMs on these datasets.
However, recent evidence reveals limitations in this paradigm.
\citet{zengOrderMattersInvestigate2025} shows that LLMs frequently exhibit neglect of some sub-context, such as each constraint to be followed,
during the attention process,
correlating with measurable performance degradation (Figure~\ref{fig:forget}.b).
This suggests that standard SFT may inadequately capture the relationship between input context and desired output,
highlighting the need for novel training strategies that more effectively leverage complex instruction datasets.

In this work, we aim to address the challenge of sub-context neglect for enhancing SFT effectiveness on complex instruction-following fine-tuning.
We start investigation with multi-constraint instruction-following \citep{qiConstraintBacktranslationImproves2024, liRuleRImprovingLLM2024},
where the model output needs to align with parallel constraints within the instruction.
We observe that in the multi-constraint scenario, reducing the number of constraints does not disrupt the alignment between the instruction and the desired output.
This suggests a potential approach to incorporate single-constraint sub-context datas for training,
which are easier for the model to learn from and eould benefit the training process.
However, single-constraint datas are generally lower-quality compared to their multi-constraint counterparts and provide limited benefits for improving complex instruction-following.

Therefore, carefully designing an effective training strategy that leverages both the full multi-constraint information and sub-context single-constraint influences is needed.
To address this modeling requirement,
we explore methods that can process multiple input sequences while building upon the capabilities of the base LLM.
We draw inspiration from the success in previous work FiD \citep{DBLP:conf/eacl/IzacardG21},
which was proposed in the retrieval-augmented generation (RAG) domain and extend encoder-decoder transformer architecture to adopt multi-input.
FiD splits the full retrieved documents into multiple inputs, each containing a single document,
and applies the modeling strategy of ``independently processing inputs and concatenating representations for the output’s attention''.
FiD provides a mixture-of-contexts paradigm that allows keeping the full instruction-output alignment
while highlighting specific sub-contexts by placing them as separate input sequences and mixing their contributions.

We investigate whether this approach is applicable to complex instruction-following fine-tuning scenarios with decoder-only transformer base LLMs.
Based on the above motivations, the key design considerations for our method include:
(1) How to structure complex instruction-following data into multiple input sequences that highlight important sub-contexts;
(2) How to achieve a mixture-of-contexts design that supports multi-sequence input; and
(3) Ensuring the modeling architecture effectively utilizes the capabilities of the base LLM.

For input structuring, we first focus on instructions with multiple parallel constraints.
Specifically, we propose including the full instruction in one input sequence,
while placing each single-constraint instruction in separate auxiliary sequences (``Parallel'' in Figure~\ref{fig:forget}.c).
This assumes the full instruction can be rewritten into its constituent sub-instructions.
For more general use cases without such assumptions, we suggest splitting the input into successive blocks, each treated as a standalone input sequence,
while maintaining full instruction-output alignment through successive position IDs (``Successive'' in Figure~\ref{fig:forget}.c).

For the modeling strategy that supports multi-sequence input,
we propose MISO, which extends current dominant decoder-only transformers to \textbf{M}ulti-\textbf{I}nput-\textbf{S}ingle-\textbf{O}utput structure.
MISO processes each input sequence independently and fuses their representations in the causal-attention layer during output generation.
Empirically, MISO fine-tuned LLMs are generally capable for standard inference,
demonstrating the feasibility of MISO as a fine-tuning method that effectively utilizes base LLMs.
We will argue that, both theoretically (section~\ref{sec-revisit}) and empirically (Appendix~\ref{sec:limitation-wo-miso}),
despite previous attempts exist \citep{ratnerParallelContextWindows2023, mcilroy-youngSetBasedPromptingProvably2024},
the design of ``independently processing inputs and concatenating representations for the output’s attention'' is unsuitable for decoder-only transformers.
Specifically, introducing more input sequences dilutes the attention allocated to the output part,
which can lead to broken generation.
To address this issue, MISO proposes to adopt ``weighted sum of attention outputs correspopnding with each input part'' recipe.

Our contributions are summarized as follows,
\begin{itemize}[leftmargin=*]
  \item We propose a novel strategy that rewrites the input sequence into multiple input sequences containing sub-contexts,
        enabling more effective fine-tuning on complex instruction-following datasets, including those involving multiple constraints.
  \item We propose MISO (Multi-Input-Single-Output), a method that extends decoder-only transformers to support multi-input modeling structures.
        MISO addresses the broken generation limitations of previous works  \citep{ratnerParallelContextWindows2023, mcilroy-youngSetBasedPromptingProvably2024} that at similar goals.
  \item Empirical results demonstrate that MISO fine-tuning offers advantages on complex instruction-following datasets,
        including settings involving multiple constraints,
        both in terms of fine-tuning effectiveness and potential training efficiency.
\end{itemize}

\section{Related Works}
\subsection{SFT for Instruction Following}
Instruction following involves adhering to user intentions to generate helpful responses,
which is fundamental to modern LLM applications \citep{zhangInstructionTuningLarge2024}.
Promoted by the success of ChatGPT \citep{ChatGPT}, performing SFT on instruction-output pairs is widely adopted
as a key step for enhancing LLM's instruction following capability \citep{ouyangTrainingLanguageModels2022, wangSelfInstructAligningLanguage2023a, li-etal-2024-quantity, liu2024what}.
The output response in the SFT datasets is generally either manual crafted \citep{ouyangTrainingLanguageModels2022} or distilled from LLMs \citep{wangSelfInstructAligningLanguage2023a}.

To meet real-world demands, instructions provided to LLMs are becoming increasingly complex.
Recent efforts to enhance complex instruction following via SFT
primarily focus on generating high-quality instruction-response pairs
through LLM-in-the-loop pipelines built upon seed instruction-tuning datasets.
WizardLM \citep{xuWizardLMEmpoweringLarge2023} rewrites the seed instruction with bread-and-depth evolution and generate response via LLMs.
Conifer \citep{sunConiferImprovingComplex2024} and RuleR \citep{liRuleRImprovingLLM2024} introduce constraint generation and response reformulation pipelines to synthesize multi-contraint instruction-output pairs.
Crab \citep{qiConstraintBacktranslationImproves2024} retains high-quality responses from seed datasets
by avoiding direct response rewriting, instead employing a constraint backtranslation technique.

Benefited from the recent success of reasoning LLMs including DeepSeek-R1 \citep{deepseek-aiDeepSeekR1IncentivizingReasoning2025},
a special way to generate complex instruction following SFT datasets is to distill reasoning texts,
which is generally thorough and complex, and capable as complex instruction.

\subsection{Multi-Input-Single-Output Transformers}
Extending transformers to support multi-input single-output architectures has attracted interest in the literature.
FiD \citep{DBLP:conf/eacl/IzacardG21} structures Retrieval-Augmented Generation (RAG) inputs as one-document-per-input,
encodes each document independently and fuses them at the cross-attention layer in encoder-decoder transformers.
PWC \citep{ratnerParallelContextWindows2023} and set-prompt \citep{mcilroy-youngSetBasedPromptingProvably2024}
adopt a similar approach with FiD, independently processing inputs and concatenating representations for attention computation in decoder-only transformers.
Our work inherits a similar design to PWC and set-prompt,
but addresses their limitation of broken generation (Section~\ref{sec-revisit}).

We note that the multi-input structure of MISO shares similarities with chunked attention mechanisms
 \citep{anTrainingFreeLongContextScaling2024, luMoBAMixtureBlock2025}.
These approaches primarily aim to simulate full attention through sparse attention,
with the goal of improving computational efficiency.
In contrast, our work fundamentally modifies the standard full attention mechanism to improve fine-tuning effectiveness.

\section{Prelimilaries}
\subsection{Multi-Constraint Instruction Following}
In this work, we aim to increase the effectiveness for SFT on complex instruction following datasets.
The term ``complex'' here refers to those instructions that are long and
contain multiple clauses, and impose several conditions that influence the expected output.
Among those, we begin our investigation with a specific scenario, multi-constraint instruction following.
Specifically, a multi-constraint instruction following dataset consists of instances of the form
$D\{x, \{c_{1}, \dots, c_{n}\}, y\}$,
where $x$ is the instruction describing the main intent,
$y$ is the output that satisfies the intent described by $x$,
$c_{1}, \dots c_{n}$ are $n$ constraints that the output $y$ need to satisfy.
We focus on the case of multi-constraint instruction following under the assumption of parallel constraints,
where each constraint $c_{1}, \dots c_{n}$ makes an independent influence on the output $y$.
Under this setting, any subset of constraints $C\subseteq\{c_{1}, \dots, c_{n}\}$
can be used to construct a valid sub-sample $D_{C}=\{x, C, y\}$.
\subsection{QKV Attention in Transformers}
We will take commonly used QKV attention (Scaled Dot-Product Attention) based transformer  \citep{vaswaniAttentionAllYou2017} for discussion in this work.
QKV attention computes weighted average of Value(V) with normalized weighting from source Query(Q) to target Key(K), formulated as,
\begin{align}
  \mathrm{Attention}(Q, K, V) = \mathrm{softmax}(\frac{QK^{T}}{\sqrt{d_{k}}})V
\end{align}
where $Q\in \mathbb{R}^{n_{src}\times d_{k}}, K\in \mathbb{R}^{n_{tgt}\times d_{k}}, V\in \mathbb{R}^{n_{tgt}\times d_{v}}$,
$n_{src/tgt}$ denotes length of source/target tokens, $d_{k}, d_{v}$ represents dimension for QK and V.

To write the KV in the chunked form, QKV attention can be written to the weighted average of attention to each KV.
\begin{align}
  &\mathrm{Attention}(Q, [K_{1}, \dots, K_{n}],[V_{1}, \dots, V_{n}])
  = \sum_{i}(\frac{S_{i}}{\sum_{i}S_{i}}\mathrm{Attention}(Q, K_{i},V_{i}))\label{eq:chunk-attention}
\end{align}
where $S_{i}=\sum \mathrm{exp}(\frac{QK_{i}^{T}}{\sqrt{d_{k}}})$ is the exponentiated attention scores from Q to $K_{i}$,
sumed over tokens.
\section{Method}
\label{sec:method}
In this section, we present MISO, a multi-input-single-output extension for decoder-only transformers.
We will first introduce the overall modeling structure of MISO (Section~\ref{sec:fio-modeling}),
and then discuss the design decision for input structuring and their application to complex instruction following fine-tuning (Section~\ref{sec:fio-data}).
We will show the recipe that leads to the design of MISO modeling, and how it overcomes the broken generation limitation of previous works  \citep{ratnerParallelContextWindows2023, mcilroy-youngSetBasedPromptingProvably2024} (Section~\ref{sec-revisit}).
\subsection{MISO Modeling Structure}
\label{sec:fio-modeling}
\begin{figure*}[t]
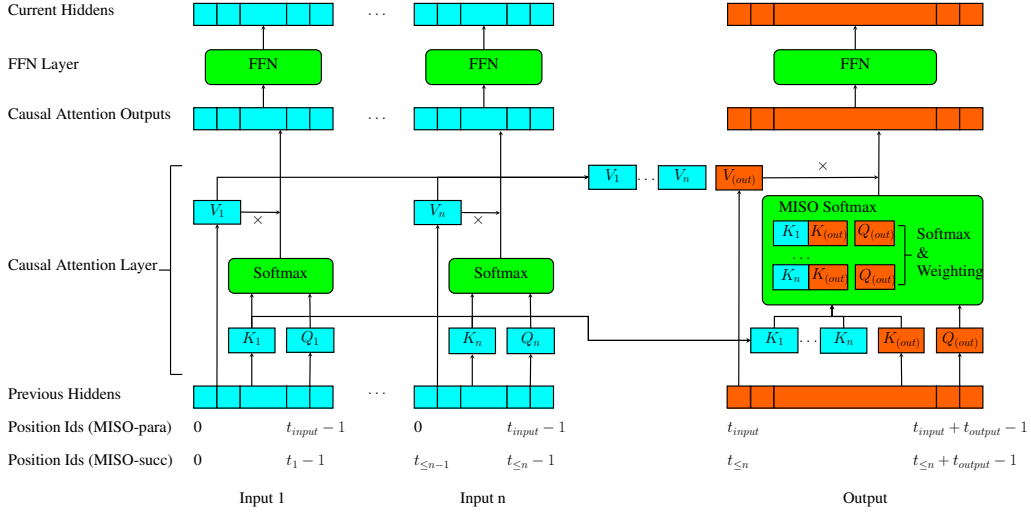

  \centering
  \resizebox{0.98\textwidth}{!}{
    \begingroup
    \huge
\ifx\du\undefined
  \newlength{\du}
\fi
\setlength{\du}{15\unitlength}


    \endgroup
  }
  \caption{Overall structure of MISO as a extension of typical transformer decoder block. Position ids for both parallel input (MISO-para) and successive input (MISO-succ) settings are illustrated.}
  \label{fig:fio}
\end{figure*}

The overall structure of MISO is illustrated in Figure~\ref{fig:fio}.
MISO adopts multiple sequences as input and performs a two-stage modeling process.
In the first stage, each input part sequence is processed independently.
In the second stage, the output part is processed using MISO’s causal-attention,
which jointly attends to all input part sequences and the output part.
The key modifications from a standard decoder-only transformer structure to MISO
include assignment of positional IDs and the attention computation for output part,
detailed as follows.

\paragraph{Position Ids}
The core design behind MISO is to enable a mixture-of-contexts paradigm with multi-input-single-output structure.
To ensure the effectiveness of mixturing, we suggest that the position id setting for each input-output pair is reasonable.
To meet this design, we provide basic MISO variants include,
1) \textbf{MISO-para}: parallel structured input sequences, which restart the positional ids from zero for each input sequence;
2) \textbf{MISO-succ}: successive structured input sequences, which successively assign position ids for successive full input sequence.

\paragraph{MISO Causal Attention}
We propose the recipe of ``weighted sum of attention outputs corresponding with each input part'' for MISO attention layer.
Specifically, to compute MISO attention from output part query $Q_{(out)}$ to both input and output part key-values
$[K_{1},V_{1},\dots,K_{n},V_{n},K_{(out)},V_{(out)}]$,
we formulate MISO causal-attention as follows,
\begin{align}
  &\mathrm{MISO\_CausalAttention}(Q_{(out)}, [K_{1}, \dots, K_{n}, K_{(out)}], [V_{1}, \dots, V_{n}, V_{(out)}])\notag \\
  =&\sum_{i} Score_{i} \cdot \mathrm{CausalAttention}(Q_{(out)}, [K_{i}, K_{(out)}], [V_{i}, V_{(out)}])\label{eq:miso}
\end{align}
where $Score_{i}$ represents a normalized weight derived from a weighting function applied to input $i$, $\sum_{i}Score_{i}=1$.

\subsection{MISO for Complex Instruction Following}
\label{sec:fio-data}
The motivation of using MISO for complex instruction following fine-tuning
is to leverage both full instruction-output correspondence and the influence of sub-context.
We consider both the specific case for multi-constraint instruction following and more general case that introduce no assumption for instruction.
The design choices for applying MISO in these scenarios are outlined as follows.
\paragraph{Inputs and Position Ids}
For multi-constraint instruction following data,
after splitting constraints from the full instruction,
we would construct single-constraint data pairs to serve as auxiliary sub-context input,
and MISO-para is then applied for this type of input.

For more general instruction cases,
rewriting the original input instruction to instructions with sub-context is not always feasible.
Instead, we split the full instruction into chunks, and treat them as multiple input sequences,
MISO-succ is then applied.

\paragraph{MISO Causal Attention}
With the motivation of highlighting the influence of sub-context,
we adopt the variant in which the weighting of input sequences is not differentiated as the default setting in MISO.
Specifically, we use $Score_{i}=\frac{1}{n}$ for every $n$ input sequences for MISO formulated in eq.\ref{eq:miso}.

\subsection{Design Behind MISO Causal-Attention}
\label{sec-revisit}
In this section, we will discuss the theoretical limitations of previous multi-input-single-output extensions for
decoder-only transformers \citep{mcilroy-youngSetBasedPromptingProvably2024},
which behave as the risk of broken generation.
Then we will revisit the previous success of multi-input-single-output extending for encoder-decoder transformer \citep{DBLP:conf/eacl/IzacardG21},
and introduce the key design principles that lead to MISO.

For multi-input-single-output extension,
the input parts are processed indepedently,
and the output part involves an attention of query from output $Q_{(out)}$
with all inputs and output part key-values $[K_{1},V_{1},\dots,K_{n},V_{n},K_{(out)},V_{(out)}]$.
In previous works \citep{ratnerParallelContextWindows2023, mcilroy-youngSetBasedPromptingProvably2024},
vanilla causal attention is applied to compute attention for output part.
\begin{align}
  &\mathrm{CausalAttention}(Q_{(out)}, [K_{1}, \dots, K_{n}, K_{(out)}],[V_{1}, \dots, V_{n}, V_{(out)}]) \notag \\
        = & \sum_{i}(\frac{S_{i}}{\sum_{i}S_{i}+S_{(out)}}\mathrm{CausalAttention}(Q_{(out)}, K_{i},V_{i})) \notag\\
  &+ \frac{S_{(out)}}{\sum_{i}S_{i}+S_{(out)}}\mathrm{CausalAttention}(Q_{(out)}, K_{(out)},V_{(out)})
\end{align}
Here we examine the ratio of attention weights allocated to output part.
For MISO-para and previous works, we structure the data so that the output is successive with all i-th input,
suppose only the i-th input exists, we have$A_{(out)}^{i} = \frac{S_{(out)}}{S_{i}+S_{(out)}}$.
The general behaviour of LLM is demonstrated to have $A_{(out)}^{i}$ siginificately large than 0,
which is crucial for effective language modeling and ensuring fluent generation \citep{xiaoEfficientStreamingLanguage2023}.
Now consider that with $n$ inputs,
\begin{align}
  A_{(out)}
  ~=~&\frac{S_{(out)}}{\sum_{i}S_{i}+S_{(out)}}~=~\frac{S_{(out)}}{\frac{\sum_{i} S_{i}}{n}+S_{(out)}}
     \frac{\frac{\sum_{i} S_{i}}{n}+S_{(out)}}{\sum_{i}S_{i}+S_{(out)}}\\
  ~=~&\mathrm{harmonic\_mean}_{i}(A_{(out)}^{i})\frac{\frac{1}{n} +\frac{S_{(out)}}{\sum_{i} S_{i}}}{1+\frac{S_{(out)}}{\sum_{i} S_{i}}}
\end{align}
when $n$ grows, in asymptotic case, $A_{(out)} \to [\mathrm{harmonic\_mean}_{i}(A_{(out)}^{i})\cdot\frac{0+0}{1+0}]$ converges to $0$.
Therefore, we argue that applying vanilla attention in the multi-input-single-output extension of a decoder-only transformer
leads to the risk of broken generation due to the diminishing attention weights on the output part.
Practical results also show that even with a moderate $n$, the generation brokes (Appendix~\ref{sec:limitation-wo-miso}).

To overcome this issue,
we rethink the success of ``independently processing inputs and concatenating representations for the output’s attention'' design
in FiD  \citep{DBLP:conf/eacl/IzacardG21}, which is based on encoder-decoder transformer architecture.
In FiD, the cross-attention for output part, where the fusion is performed, only calculate attention with input part,
which shares the formulation with eq.\ref{eq:chunk-attention} and is not affected by the diminishing $A_{(out)}$ issue.
Therefore, we propose MISO following the recipe of ``weighted sum of attention outputs corresponding with each input part'' suggested by eq.\ref{eq:chunk-attention},
\begin{align}
  &\mathrm{MISO\_CausalAttention}(Q_{(out)}, [K_{1}, \dots, K_{n}, K_{(out)}], [V_{1}, \dots, V_{n}, V_{(out)}])\notag \\
  =&\sum_{i} \frac{S_{i}+S_{(out)}}{\sum_{i}(S_{i}+S_{(out)})} \mathrm{CausalAttention}(Q_{(out)}, [K_{i}, K_{(out)}], [V_{i}, V_{(out)}])\label{eq:fio-fid}
\end{align}
this formulation avoid the diminishing $A_{(out)}$ issue by scaling $S_{(out)}$ with n.
We further replace $\frac{S_{i}+S_{(out)}}{\sum_{i}(S_{i}+S_{(out)})}$ by $Score_{i}$ for flexibility.

\section{Experiments}
\subsection{Multi-Constraint Instruction Following}
\label{sec:exp-crab}
\subsubsection{Experimental Setup}
\paragraph{Datasets and Tasks}
To demonstrate the effectiveness of MISO fine-tuning on multi-constraint instruction following tasks,
we use Crab  \citep{qiConstraintBacktranslationImproves2024} dataset for finetuning,
which is a recently proposed multi-constraint instruction following dataset with 13500 instances.
Following \citet{qiConstraintBacktranslationImproves2024}, we include ShareGPT \citep{vicuna2023} for fine-tuning.

For evaluating, we use the widely adopted IFEval dataset  \citep{zhouInstructionFollowingEvaluationLarge2023},
which is designed for multi-constraint instruction-following benchmarking.
Further, we manually flatten the IFEval dataset by decomposing each n-constraint example into n separate single-constraint instances.
The resulting dataset, named IFEval-Flatten, allows a more thoroughly evaluation for model performance on single-constraint setting besides the full multi-constraint setting.

\paragraph{Base Model and Hyperparameters}

We follow the base model and hyperparameter setting in \citet{qiConstraintBacktranslationImproves2024}.
Specifically, we use LLaMA-3-8b \citep{dubey2024llama3herdmodels} and Mistral-7b-v0.3 \citep{jiangMistral7B2023} as
base model.
Hyperparamters settings are listed in Appendix~\ref{sec:detail-crab}.
\paragraph{Comparison Models}
Our baselines include: 1) Standard SFT;
2) general instruction following LLMs,
including Vicuna-v1.5 \citep{vicuna2023}, Zephyr-7B \citep{tunstallZephyrDirectDistillation2023};
3) LLMs fine-tuned with complex instruction-following datasets except Crab,
including WizardLM-V1.2 \citep{xuWizardLMEmpoweringLarge2023},
and the Conifer  \citep{sunConiferImprovingComplex2024}.
Details are listed in Appendix~\ref{sec:detail-crab}.

For our proposed MISO,
we include both the parallel (MISO-para) and successive (MISO-succ) setting,
where the weighting across input sequences is uniformly assigned.
For MISO-para, we incorporate full multi-constraint instruction, single-constraint instructions, non-constraint instruction as input sequences.
For MISO-succ, we uniformly split instruction into 1-4 chunks for each instance.

\begin{table*}
  \centering
  \begin{tabular}{llcccccccc}
    \hline
    \multicolumn{2}{c}{} & \multicolumn{5}{c}{IFEval} & \multicolumn{3}{c}{IFEval-Flatten} \\
    Model & Base Model & [S]P & [S]I & [L]P & [L]I &Avg. & [S]I & [L]I & Avg. \\
    \hline
    \multicolumn{10}{c}{\textbf{\textit{Reference Models}}} \\
    Vicuna-v1.5* & Llama2-13B & 43.1 & 53.6 & 46.6 & 58.0 & 50.3 & - & - & - \\
    WizardLM-v1.2* & Llama2-13B & 43.6 & 54.4 & 48.4 & 59.1 & 51.4 & - & - & - \\
    Conifer-SFT* & Llama2-13B & 42.9 & 53.0 & 47.5 & 57.4 & 50.2 & - & - & - \\
    Llama3-8B* & Llama3-8B & 25.7 & 36.8 & 28.1 & 35.1 & 31.4 & - & - & - \\
    Mistral-7B* & Mistral-7B & 18.5 & 30.8 & 19.6 & 31.9 & 25.2 & - & - & - \\
    Zephyr-beta* & Mistral-7B & 32.0 & 46.8 & 44.9 & 58.0 & 45.4 & - & - & - \\
    \hline
    \multicolumn{10}{c}{\textbf{\textit{Crab Fine-tuned Models, Including Ours MISO}}} \\
    SFT & Llama3-8B & 40.1 & 51.9 & 44.7 & 55.6 & 48.1 & 52.5 & 56.5 & 54.5 \\
    \textbf{MISO-para} & Llama3-8B & 44.5 & 54.7 & \textbf{49.5} & \textbf{59.5} &  \textbf{52.1}    &     \textbf{58.4} & \textbf{62.1} & \textbf{60.3} \\
    \textbf{MISO-succ} & Llama3-8B & \textbf{44.7} & \textbf{54.9} & 47.9 & 57.9 &  51.4    &     55.4 & 58.4 & 56.9 \\
    \hline
    SFT & Mistral-7B & 38.8 & 49.4 & 41.8 & 52.5 & 45.5 & 54.0 & 57.9 & 56.0 \\
    \textbf{MISO-para} & Mistral-7B & \textbf{42.9} & \textbf{54.1} & 46.0 & 57.7 & 50.2 & \textbf{57.6} & \textbf{61.1} & \textbf{59.4} \\
    \textbf{MISO-succ} & Mistral-7B & \textbf{42.9} & 53.5 & \textbf{47.0} & \textbf{58.2} & \textbf{50.4} & 56.1 & \textbf{61.1} & 58.6 \\
    \hline
  \end{tabular}
  \caption{\label{table:crab}
    Evaluation results (\%) of the LLMs on IFEval and IFEval-Flatten,
    where ``[S]'' and ``[L]'' denote strict and loose accuracy, ``P'' and ``I'' indicate the prompt and instruction level.
    \textbf{Bold} highlights both the best result among LLMs with same base model. * denoted results from \citet{qiConstraintBacktranslationImproves2024}.
  }
\end{table*}

\subsubsection{Main Results}
We report the main results in Table~\ref{table:crab}.
The results show that MISO-para consistently outperforms the SFT baseline on both base models by a significant margin,
demonstrating the effectiveness of MISO fine-tuning compared to standard SFT.
For MISO-succ, the results also outperform SFT, yet slightly inferior to MISO-para.
The fact that the MISO fine-tuned model is capable of performing standard LLM inference suggests that the ``weighted sum of attention outputs from each input part'' design
effectively supports MISO as an extention for decoder-only transformers.

\subsubsection{Ablation Study}
For ablation study,
we evaluate an alternative sequence weighting strategy that is more similar to the vanilla attention used in FiD \citep{DBLP:conf/eacl/IzacardG21},
formulated in Eq.\ref{eq:fio-fid} and referred to as MISO-fid.

We test a version of MISO-para that excludes full multi-constraint instructions, denoted as ``- w/o full''.
Additionally, to assess whether performance improvements arise from the added single-constraint data or from the mixture-of-contexts design of MISO,
we include a control setting where we perform SFT using only the augmented data, without applying MISO, denoted as ``- no MISO''.

Results are reported in Table~\ref{table:crab-ablation},
and our observations and analysis are presented below.

\paragraph{Effectiveness of MISO fine-tuning}
MISO-para differs from SFT in these aspects:
1) it incorporates additional single-constraint data samples, and
2) it employs a mixture-of-contexts modeling structure that jointly considers these inputs.
To disentangle the contributions of these aspects, we compare against the ``- no MISO'' setting, where the same augmented data is used under standard SFT.
The results show a large performance drop in the ``- no MISO'' setting compared to both MISO-para and SFT alone.
This indicates that the single-constraint data samples are not sufficient for effective fine-tuning when used in isolation,
and that the mixture-of-contexts paradigm introduced by MISO plays a crucial role in enabling effective learning.

\begin{table*}
  \centering
  \begin{tabular}{llcccccccc}
    \hline
    \multicolumn{2}{c}{} & \multicolumn{5}{c}{IFEval} & \multicolumn{3}{c}{IFEval-Flatten} \\
    Model & Base Model & [S]P & [S]I & [L]P & [L]I &Avg. & [S]I & [L]I & Avg. \\
    \hline
    MISO-para & Llama3-8B & \textbf{44.5} & 54.7 & \textbf{49.5} & \textbf{59.5} &  \textbf{52.1}    &     \textbf{58.4} & \textbf{62.1} & \textbf{60.3} \\
    ~~- w.o. full & Llama3-8B & 44.4 & \textbf{55.4} & 48.6 & 59.4 &    52.0  &     57.3 & 60.5 & 58.9 \\
    ~~- no MISO & Llama3-8B & 38.6 & 48.7 & 41.2 & 51.7 & 45.1 & 50.5 & 53.1 & 51.8 \\
    MISO-fid-para & Llama3-8B &   \textbf{42.7} & 51.9 & \textbf{47.1} & \textbf{57.0} &   \textbf{49.7}   &     56.5 & 59.6 & 58.1 \\
    ~~- w.o. full & Llama3-8B &   41.2 & \textbf{52.5} & 44.4 & 55.4 &  48.4    &     \textbf{57.8} & \textbf{60.8} &\textbf{59.3}\\
    \hline
  \end{tabular}
  \caption{\label{table:crab-ablation}
    Ablation study results (\%) of the LLMs on IFEval and IFEval-Flatten,
    \textbf{Bold} denotes the best result with same MISO weighting variant.
  }
\end{table*}

\paragraph{Influence of MISO Weighting Strategy}
The weighting strategy is the key factor for balancing the strength of fusion,
we thus investigate whether the weighting strategy with different characteristics influences the fine-tuning effectiveness.
In addition to the default uniform weighting used in MISO-para, we evaluate MISO-fid-para,
which assigns weights based on the sum of unnormalized attention scores,
a method closer to standard attention mechanisms and biased toward longer sequences.
Ablation results show that MISO-para outperforms MISO-fid-para on both IFEval and IFEval-Flatten,
suggesting that a more balanced fusion of sub-contexts benefit the fine-tuning effectiveness for constraint following.

\paragraph{Influence of Full Multi-Constraint Input}
To investigate whether the inclusion of full multi-constraint instructions is beneficial,
we evaluate the ``- w/o full'' setting that remove the full multi-constraint input.
Results show that performance consistently drops on this setting,
indicating that incorporating full multi-constraint input during training is generally helpful.

While for MISO-fid-para is the margin between with and without full input is relatively large, for MISO-para it is slight,
indicating that with weighting strategy that makes sufficient fusion for each input,
fine-tuning effectiveness for complex multi-constraint instruction following with only single-constraint data is supported.

We note that when removing full multi-constraint input,
those single-constraint datas exactly match the single-constraint testing.
Thus, we may expect that ``- w.o. full'' ablation may performs better than full MISO-para on IFEval-Flatten,
and it is indeed the case in MISO-fid-para.
However, in MISO-para, retaining full instructions still leads to superior results.
We attribute this to the uniform weighting strategy, which allows the model to effectively learn from both full and sub-context instructions,
leading to overall better constraint-following performance.

In conclusion, while the inclusion of full multi-constraint instructions is generally beneficial,
its advantages are amplified when combined with the uniform MISO weighting strategy.

\subsection{General Complex Instruction Following}
\label{sec:exp-complex}
In Section~\ref{sec:exp-crab}, we conducted an experiment on a specific scenario involving complex instruction following with multiple parallel constraints.
The results show that both the more specific MISO-para and the more general MISO-succ outperform SFT fine-tuning.
It is worth noting that MISO cuts off the interaction between inputs, independently considers the contribution of each context to the output, and then fuses them.
We further interested in whether MISO-succ still demonstrates a fine-tuning advantage in more general and complex cases where the input instructions are densely correlated.

\subsubsection{Experimental Setup}
\paragraph{Datasets and Tasks}
Recent successes suggest a method for distilling ``instruction-think-response'' data from existing reasoning LLMs, such as DeepSeek-R1 \citep{deepseek-aiDeepSeekR1IncentivizingReasoning2025},
to enhance the reasoning capabilities of base LLMs.
We recognize the complexity of the ``thinking'' part and treat the ``instruction-think'' pair as a unified input instruction,
and fine-tuning base LLMs to generate ``response'' that align with this input.
We take 12.8K ``instruction-think-response'' samples from an existing public avaiable DeepSeek-R1-Distill dataset \footnote{https://huggingface.co/datasets/tuanha1305/DeepSeek-R1-Distill/} for fine-tuning.

\paragraph{Base Model and Hyperparameters}

We use LLaMA-3-8b \citep{dubey2024llama3herdmodels} as base model,
hyperparamters settings are listed in Appendix~\ref{sec:detail-complex}.

\begin{table*}
  \centering
  \resizebox{0.98\textwidth}{!}{
  \begin{tabular}{lccccccccccc}
    \hline
    \multirow{2}{*}{Category} & \multirow{2}{*}{And} & \multicolumn{3}{c}{\multirow{2}{*}{Chain}} & \multicolumn{4}{c}{\multirow{2}{*}{Selection}} & \multicolumn{2}{c}{Selection} & \multirow{2}{*}{All} \\
     &  & \multicolumn{3}{c}{} & \multicolumn{4}{c}{} & \multicolumn{2}{c}{\& Chain} &  \\
    \hline
    Nesting Depth & 1 & 1 & 2 & Avg. & 1 & 2 & $\geq$3 & Avg. & 2 & $\geq$3 & Avg. \\
    \hline
    Base & 23.0 & 10.7 & 13.4 & 12.7 & 12.4 & 12.2 & 16.2 & 12.2 & 7.6 & 10.6 & 16.3 \\
    SFT & 60.9 & 36.1 & 40.0 & 39.1 & 32.6 & \textbf{40.6} & 37.7 & 36.3 & \textbf{42.7} & 22.9 & 46.0 \\
    MISO-succ & \textbf{64.5} & \textbf{39.3} & \textbf{42.9} & \textbf{42.1} & \textbf{37.3} & 39.2 & \textbf{38.1} & \textbf{36.5} & 42.0 & \textbf{24.4} & \textbf{48.2} \\
    \hline
  \end{tabular}
  }
  \caption{\label{table:complex}
    Evaluation results (\%) of the compare methods on ComplexBench.
    \textbf{Bold} highlights the best result.
  }
\end{table*}

\subsubsection{Results and Discussions}
\paragraph{Effectiveness of MISO Fine-tuning}
We report evaluation results in Table~\ref{table:crab}.
Upon fine-tuning the dataset with complex instructions,
MISO-succ outperforms standard SFT, indicating the effectiveness of MISO in general complex instruction fine-tuning.
Among the evaluation terms, MISO-succ performs more significant in And \& Chain,
where the sub-instructions need to be jointly considered,
demonstrating the effecitiveness of MISO-cuss for highlighting sub-context influence.

\paragraph{Efficiency of MISO}
We note that MISO-succ demonstrates potential advantages in computational efficiency.
Considering an attention layer with time complexity $O(n_{i}^{2}+n_{i}n_{o}+n_{o}^{2})$, where $n_{i}, n_{o}$ denote the full input and output sequence lengths respectively,
uniform partitioning of the input into k chunks reduces the attention layer’s time complexity to $O(\frac{1}{k}n_{i}^{2}+n_{i}n_{o}+n_{o}^{2})$.
This reduction in input-related computation becomes advantageous when processing long input sequences.

\section{Conclusion and Limitations}
\label{sec:conclusion}
We present MISO (Multi-Input-Single-Output), a novel modeling framework that enhances complex instruction-following capabilities in large language models (LLMs) through improved supervised fine-tuning (SFT).
By extending decoder-only transformers to support multi-input-single-output structures, MISO enables a mixture-of-contexts paradigm that improves attention to critical sub-contexts while maintaining alignment with complete instruction-output pairs.
Crucially, MISO maintains compatibility with standard LLM inference pipelines.
Empirical results demonstrate the superiority of MISO as an LLM fine-tuning method, showing both enhanced performance on complex instruction-following scenarios and potential training efficiency gains.

As a general modeling paradigm, MISO has potential capabilities for applications beyond instruction tuning,
such as Retrieval-Augmented Generation (RAG) where context ordering proves critical.
We leave exploration of these applications for future work.

\newpage
\bibliography{neurips_2025}

\begin{thebibliography}{27}
\providecommand{\natexlab}[1]{#1}
\providecommand{\url}[1]{\texttt{#1}}
\expandafter\ifx\csname urlstyle\endcsname\relax
  \providecommand{\doi}[1]{doi: #1}\else
  \providecommand{\doi}{doi: \begingroup \urlstyle{rm}\Url}\fi

\bibitem[Wen et~al.(2024)Wen, Ke, Gu, Wu, Huang, Zhou, Li, Hu, Gao, Xu, Liu,
  Tang, Wang, and Huang]{wenBenchmarkingComplexInstructionFollowing2024a}
Bosi Wen, Pei Ke, Xiaotao Gu, Lindong Wu, Hao Huang, Jinfeng Zhou, Wenchuang
  Li, Binxin Hu, Wendy Gao, Jiaxin Xu, Yiming Liu, Jie Tang, Hongning Wang, and
  Minlie Huang.
\newblock Benchmarking complex instruction-following with multiple constraints
  composition, October 2024.

\bibitem[Xu et~al.(2023)Xu, Sun, Zheng, Geng, Zhao, Feng, Tao, Lin, and
  Jiang]{xuWizardLMEmpoweringLarge2023}
Can Xu, Qingfeng Sun, Kai Zheng, Xiubo Geng, Pu~Zhao, Jiazhan Feng, Chongyang
  Tao, Qingwei Lin, and Daxin Jiang.
\newblock Wizardlm: Empowering large pre-trained language models to follow
  complex instructions.
\newblock In \emph{The Twelfth International Conference on Learning
  Representations}, October 2023.

\bibitem[Qi et~al.(2024)Qi, Peng, Wang, Xu, Hou, and
  Li]{qiConstraintBacktranslationImproves2024}
Yunjia Qi, Hao Peng, Xiaozhi Wang, Bin Xu, Lei Hou, and Juanzi Li.
\newblock Constraint back-translation improves complex instruction following of
  large language models, October 2024.

\bibitem[Li et~al.(2024{\natexlab{a}})Li, Chen, Wang, Nguyen, Li, and
  Zhou]{liRuleRImprovingLLM2024}
Ming Li, Han Chen, Chenguang Wang, Dang Nguyen, Dianqi Li, and Tianyi Zhou.
\newblock Ruler: Improving llm controllability by rule-based data recycling,
  October 2024{\natexlab{a}}.

\bibitem[Zeng et~al.(2025)Zeng, He, Ren, Liang, Xiao, Zhou, Sun, and
  Yu]{zengOrderMattersInvestigate2025}
Jie Zeng, Qianyu He, Qingyu Ren, Jiaqing Liang, Yanghua Xiao, Weikang Zhou,
  Zeye Sun, and Fei Yu.
\newblock Order matters: Investigate the position bias in multi-constraint
  instruction following, March 2025.

\bibitem[Izacard and Grave(2021)]{DBLP:conf/eacl/IzacardG21}
Gautier Izacard and Edouard Grave.
\newblock Leveraging passage retrieval with generative models for open domain
  question answering.
\newblock In \emph{Proceedings of the 16th Conference of the European Chapter
  of the Association for Computational Linguistics: Main Volume, EACL 2021,
  Online, April 19 - 23, 2021}, pages 874--880. Association for Computational
  Linguistics, 2021.
\newblock \doi{10.18653/v1/2021.eacl-main.74}.

\bibitem[Ratner et~al.(2023)Ratner, Levine, Belinkov, Ram, Magar, Abend,
  Karpas, Shashua, {Leyton-Brown}, and
  Shoham]{ratnerParallelContextWindows2023}
Nir Ratner, Yoav Levine, Yonatan Belinkov, Ori Ram, Inbal Magar, Omri Abend,
  Ehud Karpas, Amnon Shashua, Kevin {Leyton-Brown}, and Yoav Shoham.
\newblock Parallel context windows for large language models, August 2023.

\bibitem[McIlroy-Young et~al.(2024)McIlroy-Young, Brown, Olson, Zhang, and
  Dwork]{mcilroy-youngSetBasedPromptingProvably2024}
Reid McIlroy-Young, Katrina Brown, Conlan Olson, Linjun Zhang, and Cynthia
  Dwork.
\newblock Order-independence without fine tuning.
\newblock In A.~Globerson, L.~Mackey, D.~Belgrave, A.~Fan, U.~Paquet,
  J.~Tomczak, and C.~Zhang, editors, \emph{Advances in Neural Information
  Processing Systems}, volume~37, pages 72818--72839. Curran Associates, Inc.,
  2024.
\newblock URL
  \url{https://proceedings.neurips.cc/paper_files/paper/2024/file/85529bc995777a74072ef63c05bedd30-Paper-Conference.pdf}.

\bibitem[Zhang et~al.(2024)Zhang, Dong, Li, Zhang, Sun, Wang, Li, Hu, Zhang,
  Wu, and Wang]{zhangInstructionTuningLarge2024}
Shengyu Zhang, Linfeng Dong, Xiaoya Li, Sen Zhang, Xiaofei Sun, Shuhe Wang,
  Jiwei Li, Runyi Hu, Tianwei Zhang, Fei Wu, and Guoyin Wang.
\newblock Instruction tuning for large language models: A survey, December
  2024.

\bibitem[{OpenAI}(2022)]{ChatGPT}
{OpenAI}.
\newblock Chatgpt, 2022.
\newblock URL \url{https://chat.openai.com/chat}.

\bibitem[Ouyang et~al.(2022)Ouyang, Wu, Jiang, Almeida, Wainwright, Mishkin,
  Zhang, Agarwal, Slama, Ray, Schulman, Hilton, Kelton, Miller, Simens, Askell,
  Welinder, Christiano, Leike, and Lowe]{ouyangTrainingLanguageModels2022}
Long Ouyang, Jeff Wu, Xu~Jiang, Diogo Almeida, Carroll~L. Wainwright, Pamela
  Mishkin, Chong Zhang, Sandhini Agarwal, Katarina Slama, Alex Ray, John
  Schulman, Jacob Hilton, Fraser Kelton, Luke Miller, Maddie Simens, Amanda
  Askell, Peter Welinder, Paul Christiano, Jan Leike, and Ryan Lowe.
\newblock Training language models to follow instructions with human feedback,
  March 2022.

\bibitem[Wang et~al.(2023)Wang, Kordi, Mishra, Liu, Smith, Khashabi, and
  Hajishirzi]{wangSelfInstructAligningLanguage2023a}
Yizhong Wang, Yeganeh Kordi, Swaroop Mishra, Alisa Liu, Noah~A. Smith, Daniel
  Khashabi, and Hannaneh Hajishirzi.
\newblock Self-instruct: Aligning language models with self-generated
  instructions.
\newblock In Anna Rogers, Jordan {Boyd-Graber}, and Naoaki Okazaki, editors,
  \emph{Proceedings of the 61st Annual Meeting of the Association for
  Computational Linguistics (Volume 1: Long Papers)}, pages 13484--13508,
  Toronto, Canada, July 2023. Association for Computational Linguistics.
\newblock \doi{10.18653/v1/2023.acl-long.754}.

\bibitem[Li et~al.(2024{\natexlab{b}})Li, Zhang, Li, Chen, Chen, Cheng, Wang,
  Zhou, and Xiao]{li-etal-2024-quantity}
Ming Li, Yong Zhang, Zhitao Li, Jiuhai Chen, Lichang Chen, Ning Cheng, Jianzong
  Wang, Tianyi Zhou, and Jing Xiao.
\newblock From quantity to quality: Boosting {LLM} performance with self-guided
  data selection for instruction tuning.
\newblock In Kevin Duh, Helena Gomez, and Steven Bethard, editors,
  \emph{Proceedings of the 2024 Conference of the North American Chapter of the
  Association for Computational Linguistics: Human Language Technologies
  (Volume 1: Long Papers)}, pages 7602--7635, Mexico City, Mexico, June
  2024{\natexlab{b}}. Association for Computational Linguistics.
\newblock \doi{10.18653/v1/2024.naacl-long.421}.
\newblock URL \url{https://aclanthology.org/2024.naacl-long.421/}.

\bibitem[Liu et~al.(2024)Liu, Zeng, He, Jiang, and He]{liu2024what}
Wei Liu, Weihao Zeng, Keqing He, Yong Jiang, and Junxian He.
\newblock What makes good data for alignment? a comprehensive study of
  automatic data selection in instruction tuning.
\newblock In \emph{The Twelfth International Conference on Learning
  Representations}, 2024.
\newblock URL \url{https://openreview.net/forum?id=BTKAeLqLMw}.

\bibitem[Sun et~al.(2024)Sun, Liu, Li, Wang, Dong, Lin, and
  Huang]{sunConiferImprovingComplex2024}
Haoran Sun, Lixin Liu, Junjie Li, Fengyu Wang, Baohua Dong, Ran Lin, and Ruohui
  Huang.
\newblock Conifer: Improving complex constrained instruction-following ability
  of large language models, April 2024.

\bibitem[DeepSeek-AI(2025)]{deepseek-aiDeepSeekR1IncentivizingReasoning2025}
DeepSeek-AI.
\newblock Deepseek-r1: Incentivizing reasoning capability in llms via
  reinforcement learning, January 2025.

\bibitem[An et~al.(2024)An, Huang, Zhang, Gong, Qiu, Zhou, and
  Kong]{anTrainingFreeLongContextScaling2024}
Chenxin An, Fei Huang, Jun Zhang, Shansan Gong, Xipeng Qiu, Chang Zhou, and
  Lingpeng Kong.
\newblock Training-free long-context scaling of large language models, May
  2024.

\bibitem[Lu et~al.(2025)Lu, Jiang, Liu, Du, Jiang, Hong, Liu, He, Yuan, Wang,
  Huang, Yuan, Xu, Xu, Lai, Chen, Zheng, Yan, Su, Wu, Zhang, Yang, Zhou, Zhang,
  and Qiu]{luMoBAMixtureBlock2025}
Enzhe Lu, Zhejun Jiang, Jingyuan Liu, Yulun Du, Tao Jiang, Chao Hong, Shaowei
  Liu, Weiran He, Enming Yuan, Yuzhi Wang, Zhiqi Huang, Huan Yuan, Suting Xu,
  Xinran Xu, Guokun Lai, Yanru Chen, Huabin Zheng, Junjie Yan, Jianlin Su,
  Yuxin Wu, Neo~Y. Zhang, Zhilin Yang, Xinyu Zhou, Mingxing Zhang, and Jiezhong
  Qiu.
\newblock Moba: Mixture of block attention for long-context llms, February
  2025.

\bibitem[Vaswani et~al.(2017)Vaswani, Shazeer, Parmar, Uszkoreit, Jones, Gomez,
  Kaiser, and Polosukhin]{vaswaniAttentionAllYou2017}
Ashish Vaswani, Noam Shazeer, Niki Parmar, Jakob Uszkoreit, Llion Jones,
  Aidan~N Gomez, {\L}ukasz Kaiser, and Illia Polosukhin.
\newblock Attention is all you need.
\newblock In \emph{Advances in Neural Information Processing Systems},
  volume~30. Curran Associates, Inc., 2017.

\bibitem[Xiao et~al.(2023)Xiao, Tian, Chen, Han, and
  Lewis]{xiaoEfficientStreamingLanguage2023}
Guangxuan Xiao, Yuandong Tian, Beidi Chen, Song Han, and Mike Lewis.
\newblock Efficient streaming language models with attention sinks.
\newblock In \emph{The Twelfth International Conference on Learning
  Representations}, October 2023.

\bibitem[Chiang et~al.(2023)Chiang, Li, Lin, Sheng, Wu, Zhang, Zheng, Zhuang,
  Zhuang, Gonzalez, Stoica, and Xing]{vicuna2023}
Wei-Lin Chiang, Zhuohan Li, Zi~Lin, Ying Sheng, Zhanghao Wu, Hao Zhang, Lianmin
  Zheng, Siyuan Zhuang, Yonghao Zhuang, Joseph~E. Gonzalez, Ion Stoica, and
  Eric~P. Xing.
\newblock Vicuna: An open-source chatbot impressing gpt-4 with 90\%* chatgpt
  quality, March 2023.
\newblock URL \url{https://lmsys.org/blog/2023-03-30-vicuna/}.

\bibitem[Zhou et~al.(2023)Zhou, Lu, Mishra, Brahma, Basu, Luan, Zhou, and
  Hou]{zhouInstructionFollowingEvaluationLarge2023}
Jeffrey Zhou, Tianjian Lu, Swaroop Mishra, Siddhartha Brahma, Sujoy Basu,
  Yi~Luan, Denny Zhou, and Le~Hou.
\newblock Instruction-following evaluation for large language models, November
  2023.

\bibitem[AI@Meta()]{dubey2024llama3herdmodels}
AI@Meta.
\newblock The llama 3 herd of models.

\bibitem[Jiang et~al.(2023)Jiang, Sablayrolles, Mensch, Bamford, Chaplot,
  de~las Casas, Bressand, Lengyel, Lample, Saulnier, Lavaud, Lachaux, Stock,
  Scao, Lavril, Wang, Lacroix, and Sayed]{jiangMistral7B2023}
Albert~Q. Jiang, Alexandre Sablayrolles, Arthur Mensch, Chris Bamford,
  Devendra~Singh Chaplot, Diego de~las Casas, Florian Bressand, Gianna Lengyel,
  Guillaume Lample, Lucile Saulnier, L{\'e}lio~Renard Lavaud, Marie-Anne
  Lachaux, Pierre Stock, Teven~Le Scao, Thibaut Lavril, Thomas Wang,
  Timoth{\'e}e Lacroix, and William~El Sayed.
\newblock Mistral 7b, October 2023.

\bibitem[Tunstall et~al.(2023)Tunstall, Beeching, Lambert, Rajani, Rasul,
  Belkada, Huang, von Werra, Fourrier, Habib, Sarrazin, Sanseviero, Rush, and
  Wolf]{tunstallZephyrDirectDistillation2023}
Lewis Tunstall, Edward Beeching, Nathan Lambert, Nazneen Rajani, Kashif Rasul,
  Younes Belkada, Shengyi Huang, Leandro von Werra, Cl{\'e}mentine Fourrier,
  Nathan Habib, Nathan Sarrazin, Omar Sanseviero, Alexander~M. Rush, and Thomas
  Wolf.
\newblock Zephyr: Direct distillation of lm alignment, October 2023.

\bibitem[Cui et~al.(2023)Cui, Yuan, Ding, Yao, Zhu, Ni, Xie, Liu, and
  Sun]{cuiUltraFeedbackBoostingLanguage2023}
Ganqu Cui, Lifan Yuan, Ning Ding, Guanming Yao, Wei Zhu, Yuan Ni, Guotong Xie,
  Zhiyuan Liu, and Maosong Sun.
\newblock Ultrafeedback: Boosting language models with high-quality feedback,
  October 2023.

\bibitem[Rafailov et~al.(2024)Rafailov, Sharma, Mitchell, Ermon, Manning, and
  Finn]{rafailovDirectPreferenceOptimization2024}
Rafael Rafailov, Archit Sharma, Eric Mitchell, Stefano Ermon, Christopher~D.
  Manning, and Chelsea Finn.
\newblock Direct preference optimization: Your language model is secretly a
  reward model, July 2024.

\end{thebibliography}


\appendix
\newpage
\section*{Appendix}
\section{Empirical Limitation for Extending Multi-Input-Single-Output with Vanilla Attention}
\label{sec:limitation-wo-miso}
PWC \citep{ratnerParallelContextWindows2023} and set-prompt \citep{mcilroy-youngSetBasedPromptingProvably2024}
are previous methods that adopt a similar approach with FiD \citep{DBLP:conf/eacl/IzacardG21} and our proposed MISO,
they independently processing inputs and concatenating representations for attention computation in decoder-only transformers.

Results reported in set-prompt \citep{mcilroy-youngSetBasedPromptingProvably2024} has shown performance degration under natural language understanding(NLU) evaluation with relatively larger output sequence length.
We attribute this to the accumulation of error introduced by output part attention dilution.

For case study, we take a data sample from \citet{mcilroy-youngSetBasedPromptingProvably2024},
and demonstrate the LLaMA-3-8b-Instruct \citep{dubey2024llama3herdmodels} generation results with and without MISO (Table~\ref{table:set-prompt}).
MISO variant is set to MISO-fid formulated in eq.\ref{eq:fio-fid}.

Result shows that, although may still feasible for getting right answer under NLU setting,
set-prompt failed to generate fluent response,
while MISO fix that.

\begin{table*}
  \center
  \begin{tabular}{r|l}
    \hline
    Inputs Template& A lesion causing compression of the facial nerve at the \\
    & stylomastoid foramen will cause ipsilateral \{Choice\} Answer:\\
    Choices & ``1. paralysis of the facial muscles.'' \\
                   & ``2. paralysis of the facial muscles and loss of taste.'' \\
                   & ``3. paralysis of the facial muscles, loss of taste and lacrimation.'' \\
                   & ``4. paralysis of the facial muscles, loss of taste, lacrimation\\
    & ~~~~and decreased salivation.''\\
    \hline
    set-prompt generation & 1. 2. 3. 4 \\
    MISO generation & The facial nerve is responsible for the paralysis of the\\
    &facial muscles and loss of taste.\\
    \hline
  \end{tabular}
  \caption{\label{table:set-prompt}Case for extending multi-input-single-output with vanilla attention \& MISO.}
\end{table*}

\section{Detailed Experiment Setups for Multi-Constraint Instruction Following}
\label{sec:detail-crab}
\paragraph{Comparison Models}
Details for our comparision models in multi-constraint instruction following experiment setting include,
1) general instruction following LLMs,
including Vicuna-v1.5 \citep{vicuna2023} that trained on the 125k ShareGPT dataset,
and Zephyr-7B \citep{tunstallZephyrDirectDistillation2023}, trained with the UltraFeedback \citep{cuiUltraFeedbackBoostingLanguage2023} dataset using DPO \citep{rafailovDirectPreferenceOptimization2024},
which achieves leading performance on chat benchmarks based on Mistral-7b.
2) LLMs fine-tuned with complex instruction-following datasets except Crab,
including WizardLM-V1.2 \citep{xuWizardLMEmpoweringLarge2023} that trained on the 196k evolutioned instruction tuning dataset,
and the Conifer  \citep{sunConiferImprovingComplex2024}, which is a recent work that generate multi-constraint instruction tuning data with GPT4 involved pipeline.
\paragraph{Hyperparameters}
We follow the hyperparameter setting for fine-tuning in \citet{qiConstraintBacktranslationImproves2024}.
Specifically, we set the learning rate to 5e-6,
with a micro batch of 1 and macro batch size of 256. The warm-up ratio is set to 0.1.
The Mistral-7B experiments are trained for 4 epochs with a maximum sequence length of 2048,
while the LLaMA-3-8B experiments are trained for 3 epochs with a maximum sequence length of 4096.

\paragraph{Resources}
Experiments are conducted on 8 Hygon DCU for training, with ROCm compability.
Training is conducted using Deepspeed pipeline parallelism\footnote{https://github.com/deepspeedai/DeepSpeed}.
The LLaMA-3-8B fine-tuning took around 33 hours.

The Crab \citep{qiConstraintBacktranslationImproves2024} dataset and Mistral model used are under Apache-2.0 license.

\section{Detailed Experiment Setups for General Complex Instruction Following}
\label{sec:detail-complex}
\paragraph{Hyperparameters}
We set the learning rate to 5e-6,
with a macro batch size of 128. The warm-up ratio is set to 0.1.
We fine-tune LLaMA-3-8B for 4 epochs with a maximum sequence length of 4096.
Chunk number for MISO-succ is sampled in [1..4] with probability of [0.55, 0.15, 0.15, 0.15].

\paragraph{Resources}
Same with multi-constraint instruction following experiment setting,
experiments are conducted on 8 Hygon DCU for training, with ROCm compability.
Training is conducted using Deepspeed pipeline parallelism\footnote{https://github.com/deepspeedai/DeepSpeed}.

We use qwen-plus-2025-04-28 API for evaluating on ComplexBench.

\end{document}